%% file: iclr2025_conference.tex
\documentclass{article} % For LaTeX2e
\usepackage{iclr2025_conference,times}

\input{math_commands.tex}

\usepackage{hyperref}
\usepackage{url}
\usepackage{enumitem, graphicx, multirow, amsfonts, amsmath, verbatim,algorithm, algpseudocode, booktabs, array, arydshln, times, epsfig, amssymb, pifont, capt-of, wrapfig, lipsum, makecell, tabulary, etoolbox, subcaption, color, fontawesome}
\usepackage[algo2e]{algorithm2e}

\definecolor{darkgreen}{rgb}{0.45, 0.57, 0.37}
\def\eg{\emph{e.g. }} 
\def\ie{\emph{i.e. }} 
\def\etal{\emph{et~al. }} 
\algnewcommand{\IIf}[1]{\State\algorithmicif\ #1\ \algorithmicthen}
\algnewcommand{\EndIIf}{\unskip\ \algorithmicend\ \algorithmicif}

\title{Long-horizon Visual Instruction Generation with Logic and Attribute Self-reflection}

\author{Yucheng Suo \quad Fan Ma \quad Kaixin Shen \quad Linchao Zhu \quad \textbf{Yi Yang} \thanks{Corresponding Author.}\\ 
ReLER Lab, CCAI, Zhejiang University, China  \\
}

\iclrfinalcopy % Uncomment for camera-ready version, but NOT for submission.
\begin{document}

\maketitle

\begin{figure}[h]
    \centering
    \vspace{-1\intextsep}
    \includegraphics[width=0.99\linewidth]{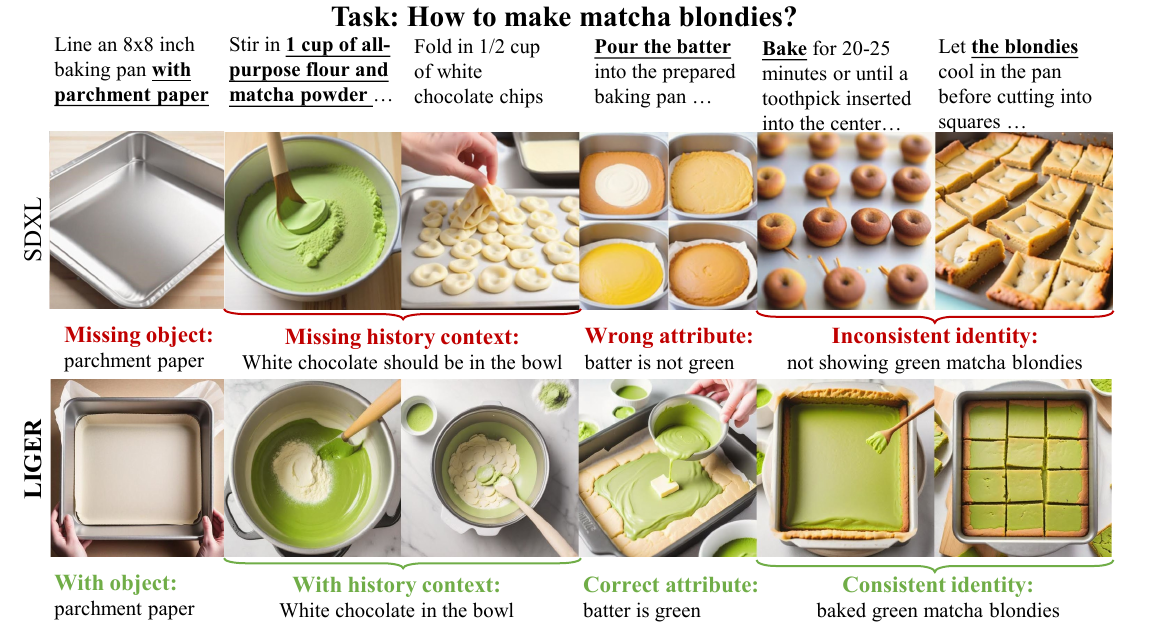}
    \vspace{-0.5\intextsep}
    \caption{Visual instruction generated by LIGER, key merits are \textcolor{darkgreen}{highlighted} in the figure.}  %\textcolor{blue}{(1) Correct object and attribute}, \textcolor{green}{(2) Adequate scene arrangement}, \textcolor{orange}{(3) object consistent between consecutive steps}. }
    %\vspace{-0\intextsep}
    \label{fig:motivation}
\end{figure}

\begin{abstract}
Visual instructions for long-horizon tasks are crucial as they intuitively clarify complex concepts and enhance retention across extended steps. 
Directly generating a series of images using text-to-image models without considering the context of previous steps results in inconsistent images, increasing cognitive load. Additionally, the generated images often miss objects or the attributes such as color, shape, and state of the objects are inaccurate.
To address these challenges, we propose \textbf{LIGER}, the first training-free framework for \textbf{L}ong-horizon \textbf{I}nstruction \textbf{GE}neration with logic and attribute self-\textbf{R}eflection. LIGER first generates a draft image for each step with the historical prompt and visual memory of previous steps. This step-by-step generation approach maintains consistency between images in long-horizon tasks. Moreover, LIGER utilizes various image editing tools to rectify errors including wrong attributes, logic errors, object redundancy, and identity inconsistency in the draft images. Through this self-reflection mechanism, LIGER improves the logic and object attribute correctness of the images.
To verify whether the generated images assist human understanding, we manually curated a new benchmark consisting of various long-horizon tasks. Human-annotated ground truth expressions reflect the human-defined criteria for how an image should appear to be illustrative. 
Experiments demonstrate the visual instructions generated by LIGER are more comprehensive compared with baseline methods. Code and dataset are provided in \url{https://github.com/suoych/LIGER}.
\end{abstract}

\section{Introduction}\label{intro}
Humans learn to accomplish real-world tasks quickly through step-by-step text instructions. However, without visual aids, it is challenging to imagine the object attribute status and judge the completion status of the steps. For instance, when frying potato chips, merely reading the text description makes it hard to judge whether the chips are done. In contrast, viewing a video or a series of images accelerates individual understanding of task procedures, enhancing the success rate of completing various tasks. Generating illustrative visual instructions eases the comprehension burden and therefore becomes a crucial and trending task \citep{lu2023multimodal,bordalo2024generating,menon2024generating,damen2024genhowto}. Moreover, generating visual instructions unleashes the potential applications including multi-modal embodied agent perception and new task adaptation \citep{fan2024bevinstructor,zhou2024image}. In this paper, we aim to generate a series of images given task step descriptions.

% describe challenge
A naive approach to generating visual instructions involves directly using text-to-image models, such as Latent Diffusion Models (LDMs) \citep{rombach2022high}. As Figure \ref{fig:motivation} illustrates, this method results in images lacking object consistency, thereby confusing users about the relationships between steps. To enhance image continuity, GenHowTo \citep{damen2024genhowto} trains a controllable U-Net \citep{ronneberger2015u} model to enhance identity consistency. StackDiffusion \citep{menon2024generating} uses a diffusion model that takes concatenated latents from different steps as input. Sequential Latent Diffusion Model (SLDM) \citep{bordalo2024generating} trains a language model to regenerate consistent textual descriptions and use latents of the previous steps to enhance consistency. However, these approaches tend to produce overly consistent images that fail to capture changes in object states. An illustrative visual instruction should balance continuity with sufficient variability. This leads to the first challenge: the need for logical coherence across steps while allowing for appropriate changes.
%The first challenge can be summarized as \textbf{Logical Inaccuracy}.

%Consider generating images for the task of \textit{How to make matcha blondies}, the following two steps: \textit{Bake for 20 minutes} and \textit{let it cool} should show blondies with a consistent appearance. 
Moreover, we empirically observe that the attributes of objects, \eg color, state, and shape, might be incorrect in the images as depicted in Figure \ref{fig:motivation}. These errors can accumulate, impacting the generation result of other steps and posing a significant challenge in long-horizon tasks. This leads to the second challenge, \ie, attribute error and cumulation.  

%Introduce previous methods limitations
%tedious training procedure requires a step-wise annotated dataset and tremendous computing resources. Moreover, a consensus limitation of these methods is that \textbf{the number of steps for a task is restricted to a fixed number}. This further limits the practicability when the users accomplish more complex tasks. Is there a way to break such limits without training, helping users with tasks of various complexity? 

Our intuition for addressing these issues is to first generate a draft image for each step with the visual and textual context of previous steps, ensuring continuity between images. Then, through a process of self-reflection, we refine the draft images by adjusting for excessive continuity and correcting object attribute errors. This iterative approach not only prevents the accumulation of attribute errors in long-horizon tasks but also maintains appropriate logic relations across steps, similar to drafting and refining sketches. %This generate-rectify paradigm reflects a self-reflection mechanism.

%Introduce our method
To this end, we propose LIGER, a training-free framework for long-horizon visual instruction generation consisting of (1) historical prompt and visual memory, (2) self-reflection and memory calibration. Specifically, we leverage the reasoning ability of LLM to explicitly output history context for each step, facilitating relation comprehension. Inspired by the recent training-free identity consistent generation works \citep{zhou2024storydiffusion, tewel2024training}, LIGER additionally injects the previous step visual latent embedding into the frozen text-to-image diffusion model, generating coherent images for different steps. 
To further refine the object attribute in the images and avoid over-consistent, 
%we introduce an tool-based self-correction mechanism exploiting the multi-modal perception and reasoning ability of the Multi-modal Large Language Model (MLLM). In detail, 
a MLLM receives multi-modal in-context prompting and tells the rectifying solutions. Various editing tools deal with errors including attribute error, object redundancy, identity inconsistency, and logic misunderstanding. Then the visual memory is calibrated to the embedding of the edited image via a latent inversion procedure, avoiding the error affecting future step image generation. Having this step-by-step generation manner, LIGER is capable of tasks with arbitrary steps without training.

To evaluate whether the generated visual instructions align with human comprehension, we curate a benchmark containing 569 long-horizon tasks along with human-annotated ground truth expressions and logic relations. Moreover, we evaluate the method from semantic alignment, logic correctness, and illustrativeness. %Moreover, we design new evaluation metrics testing semantic alignment, logic correctness, and illustrativeness. 
Results show that LIGER surpasses baseline methods by a large margin. User studies and qualitative comparisons further verify that visual instructions generated by LIGER are more illustrative.
In summary, the contribution of this paper includes:

(1) We propose LIGER, the first training-free framework generating visual instructions for long-horizon tasks. 

(2) History prompts, visual memory, and self-reflection are introduced to promise logic coherent and object property accuracy. Inversion-based memory calibration is devised to avoid exposure bias.

(3) A dataset for long-horizon tasks with human-annotated expressions is curated to evaluate the effectiveness of LIGER.
%(3) A dataset for long-horizon tasks with human-annotated expressions and a set of evaluation metrics are designed to evaluate the effectiveness of LIGER.

\section{Related Work}
\label{related}
\subsection{Image Generation and Editing}
Recent advances in multi-modal diffusion models \citep{ramesh2022hierarchical,koh2024generating,peebles2023scalable,saharia2022photorealistic,ho2020denoising,song2020score} show a remarkable ability to generate images in high fidelity. Among these models, Latent diffusion models (LDMs) \citep{rombach2022high} show strong robustness and semantic richness since the denoising process is conducted on the latent space. Based on LDMs, researchers further exploit exciting application topics including controllable image generation \citep{zhang2023adding,mou2024t2i,liang2024caphuman,ma2024directed}, personalized generation \citep{ruiz2023dreambooth,kumari2023multi,shi2024instantbooth,gal2022image}, coherent generation \cite{zhou2024storydiffusion,tewel2024training}, image editing \citep{brooks2023instructpix2pix,hertz2022prompt,nichol2021glide,kim2022diffusionclip,mou2023dragondiffusion,shi2024dragdiffusion,mou2024diffeditor}, etc. Storydiffusion \citep{zhou2024storydiffusion} and Consistory \citep{tewel2024training} share a similar idea of KV sharing to generate content-consistent images in a training-free manner. 

Image editing, different from previous image generation tasks, involves manipulating the contents of the given image \citep{pan2023drag}. There are various settings for editing, including text-driven \citep{tumanyan2023plug,cao2023masactrl,kawar2023imagic}, location-based \citep{chen2024anydoor,avrahami2023blended,nichol2021glide}, appearance modulation \citep{chen2024zero,mou2023dragondiffusion}, object moving \citep{pan2023drag,mou2024diffeditor}, etc. Common techniques for text-guided editing involve modifying the latent attention module \eg MasaCtrl \citep{cao2023masactrl} or fine-tuning a model \eg Instructpix2pix and SmartEdit \citep{brooks2023instructpix2pix,huang2024smartedit}. Location-based editing leverages the region restriction prior like bounding box, mask, or even point \citep{ling2023freedrag}. Our method utilizes different image editing methods to rectify the errors in the image. 
\subsection{Task Instruction Generation}
Generating procedures for a task is a popular research topic as it has potential application scenarios like intelligent assistants \citep{shen2024hugginggpt,suris2023vipergpt,yang2024doraemongpt}, embodied agents navigation \citep{liu2023bird, liu2024volumetric} and instruction comprehension \citep{xu2023wizardlm}, etc. This paper focuses on visual instruction generation, \ie generating a series of images to explain a task. Previous work like TIP \citep{lu2023multimodal} and MGSL \citep{wang2022multimedia} generates textual instructions for the tasks based on the visual information. StackDiffusion \citep{menon2024generating} is the first method for generating coherent visual instructions, which is trained on step-wise annotated VSGI dataset \citep{yang2021visual}. However, the step number for a task is restricted. GenHowTo \citep{damen2024genhowto} infers states before and after actions by learning from instructional videos. Sequential Latent Diffusion Model \citep{bordalo2024generating} trains a model to output coherent text prompts for the text-to-image diffusion model, therefore generating coherent images. Phung \etal  \cite{phung2024coherent} propose a training-free method yet the utmost step length is 5. Different from previous methods, LIGER is a training-free method that can deal with long-horizon tasks having large step lengths.
\subsection{Tool-based methods}
As the growing emergent capabilities of LLMs \citep{achiam2023gpt}, researchers deal with complex vision and natural language tasks \citep{yao2022react,Ma_2024_CVPR} by using surrogate tools \citep{schick2024toolformer} or programming languages, pioneer works include VisProg \citep{gupta2023visual}, ViperGPT \citep{suris2023vipergpt}, HuggingGPT \citep{shen2024hugginggpt}, etc. In the image and video generation area, LLMs are widely used for arranging layouts \citep{gani2023llm,lin2023videodirectorgpt,lian2023llm,yang2024mastering}, enriching textual prompts \citep{cheng2024autostudio,long2024videodrafter,yuan2024mora,zhuang2024vlogger}, tool calling \citep{wang2024genartist}, verification \citep{wu2024self}. Our method is also a tool-based framework unleashing the strong reasoning ability of Multi-modal Large Language Models (MLLMs) to call tools, enrich textual information, and do self-reflection.

\begin{figure}[t]
    \centering
    \vspace{-0.5\intextsep}
    \includegraphics[width=\linewidth]{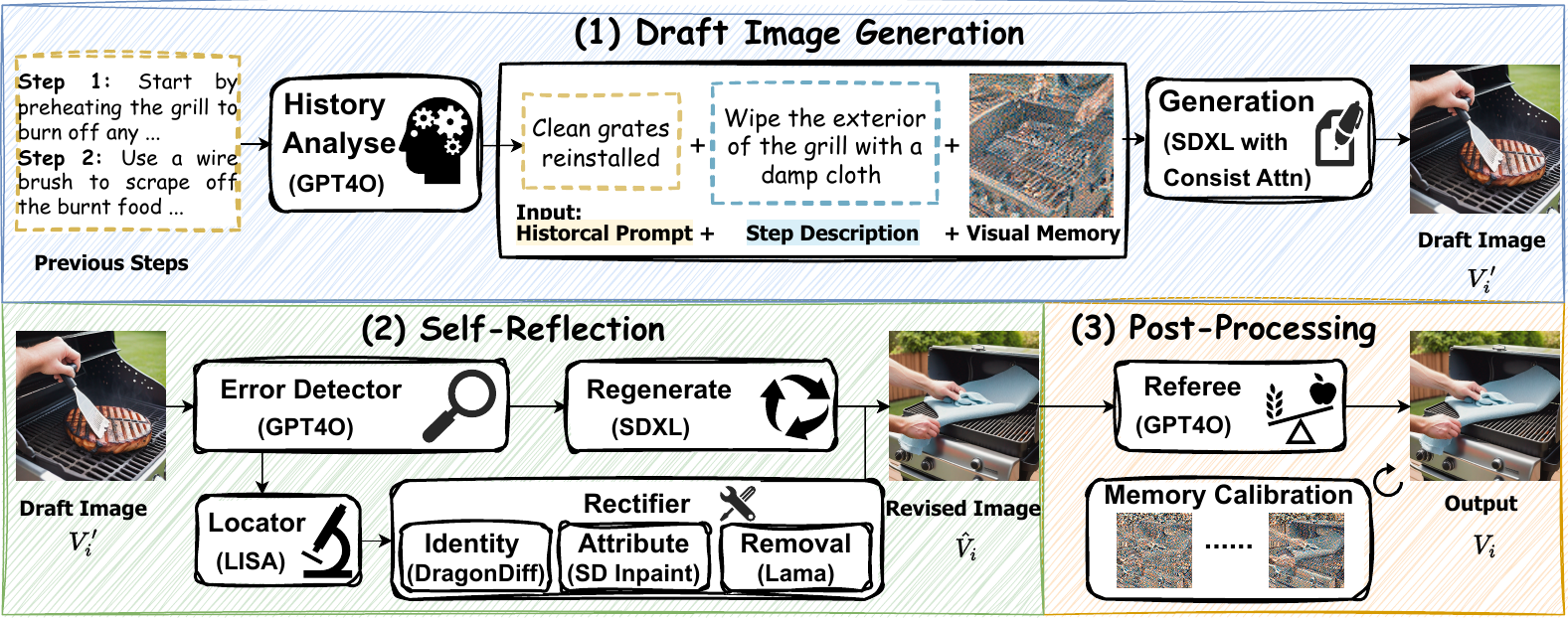}
    \vspace{-1.2\intextsep}
    \caption{\textbf{Pipeline overview.} LIGER generates visual instructions step-by-step, starting with (1) generating a draft image taking the visual memory, step description and historical prompt as input. (2) The error detector identifies the error and the corresponding tool fixes it, generating a revised image. (3) The referee tool compares the two images and selects one as the final output. LIGER further uses inversion-guided visual memory calibration for future step generation.}
    \vspace{-1\intextsep}
    \label{fig:pipeline}
\end{figure}

\section{Method}\label{headings}
The overall pipeline of LIGER is shown in Figure \ref{fig:pipeline}. Harnessing the visual memory and historical prompt, LIGER generates a draft image for each step. Self-reflection mechanism corrects the errors in the draft images. To prevent error accumulation in the long horizon generation procedure, LIGER calibrates the visual memory according to the edited image through inversion.

\subsection{History-aware Draft Image Generation}\label{history}
Given a set of step descriptions $\sS^n$ for a task $Q$ of $n$ steps, our goal is to generate a series of coherent images $\sV$ for corresponding descriptions without training. To this end, a frozen text-to-image diffusion model generates a draft image $V'_i$ for step $i$ for each step in the task. The diffusion model generates a single image through iterative denoising steps. Specifically, a U-Net network $U$ predicts the noise 
\begin{equation}
\label{eq1}
    \epsilon_t = U(\vz_t, \vc),
\end{equation}
where $\vz_t$ is the latent representation at timestep $t$ and $\vc$ is the textual condition. Naively generating individual images using the step description ignores the continuity between steps. Therefore, we first introduce the historical prompt and visual memory to enhance consistency.\\
\textbf{Historical prompt.} Each step description $S_i \in \sS$ often describes an incremental action relative to the previous scene settings. For instance, in a task \textit{cooking potato chips}, two consecutive steps are: \textit{place the potato chips on a paper towel to drain excess oil} and \textit{seasoning with salt and pepper}. Without context, the text-to-image diffusion model is unaware that salt and pepper should be added to the potato chips. Motivated by this, we use an LLM to generate a description $H_i$ for each step that specifies which objects from the previous steps should appear in the current step. The text condition $\vc$ for the diffusion model is formulated as 
\begin{equation}
\label{eq2}
    \vc = E_T(S_i, H_i),
\end{equation}
where $E_T$ is the text encoder network. \\
\textbf{Visual memory sharing.} Merely using the historical prompt results in generating objects with varied appearances and backgrounds. To address this issue, inspired by StoryDiffusion \citep{zhou2024storydiffusion}, we incorporate visual embeddings from the previous step as the visual context. When generating the draft image $V'_i$ of step $i$, we randomly sample several visual feature tokens $p_{i-1} \in \sR^{M\times C}$ of the previous image $V_{i-1} \in \sV$ and inject them into the self-attention operation in the U-Net. Here $M$ represents the number of sampled tokens and $C$ is the number of feature channels. The query input of the attention operation is the current image feature tokens $p_i \in \sR^{N\times C}$, the key and value inputs are the concatenation of $p_{i-1}$ and $p_i$. The procedure can be formulated as:
\begin{equation}\label{eq3}
\begin{aligned}
    Q_i = W^q\vp_i, K_i &= W^k[\vp_i,\vp_{i-1}], V_i = W^v[\vp_i,\vp_{i-1}], \\
    O_i &= Attention(Q_i,K_i,V_i),
\end{aligned}
\end{equation}
where $W^q$, $W^k$,$W^v$ are the linear projection layers for the query, key, and value respectively. The output feature $O_i$ is used as the input of the next layer in the UNet $U$.
Note that neither the historical prompt nor the visual memory are provided in the first step of any task. 

\begin{figure}[t]
    \centering
    \vspace{-0.5\intextsep}
    \includegraphics[width=\linewidth]{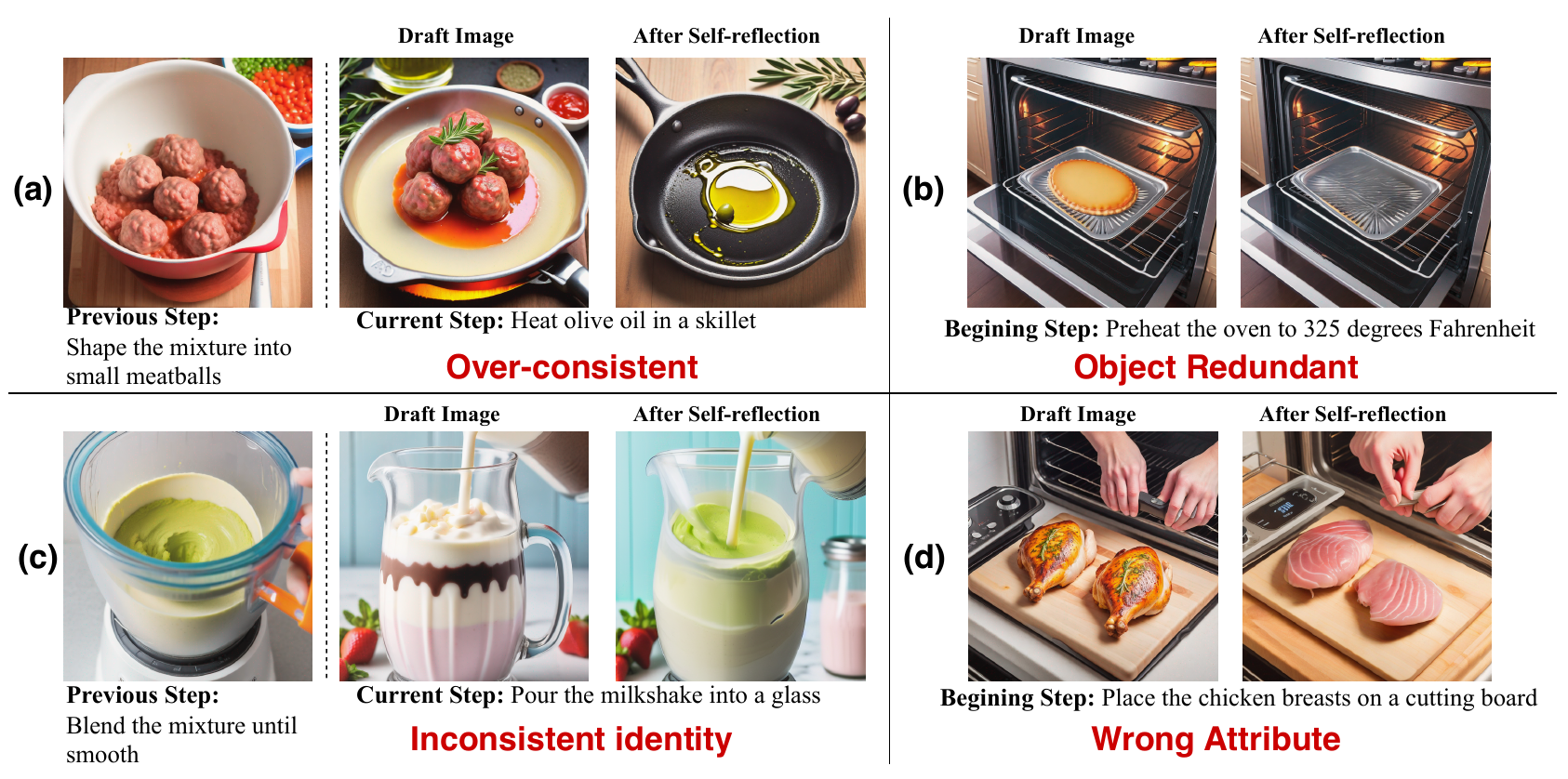}
    \vspace{-1.5\intextsep}
    \caption{Visualization of different error types and the effect of self-reflection. The motivation of self-reflection is to rectify errors including (a) over-consistent, (b) object redundant, (c) inconsistent identity, and (d) wrong attributes. }
    \vspace{-1.5\intextsep}
    \label{fig:errortype}
\end{figure}

\subsection{Tool-based Self-reflection}
Empirically, we observe errors in the draft images as illustrated in Figure \ref{fig:errortype}. Leveraging the advanced multi-modal capabilities of MLLMs, LIGER employs the state-of-the-art GPT4O model as an error detector to identify errors across four aspects, then output tool calling instructions to revise the draft images. For accuracy in error recognition, the error detector is prompted with multimodal in-context examples. The prompt template is attached in the appendix.\\
\textbf{Over-consistent.}
In long-horizon tasks, not all steps necessarily require visual continuity. For example, consider the task of \textit{cooking wanton noodles} where the steps \textit{Drain the noodles and rinse with cold water} and \textit{In a separate pan, heat some oil} are sequential yet independent. The former step concludes noodle preparation, while the latter step initiates cooking with different ingredients. 
These steps lack logistic connection, making consistency between the two images unnecessary. Breaking this consistency can help users recognize the transition to a new step. To address the over-consistent issue, the error detector assesses whether to maintain or disrupt the continuity. If breaking consistency is required, the error detector outputs the error rectification instruction in the format of $Regenerate(New \ text)$, then regenerates an image according to the new description. \\
\textbf{Identity inconsistent.} Despite historical prompt and visual memory contributing to global visual consistency, local details occasionally remain misaligned, as depicted in Figure \ref{fig:errortype}. To enhance local consistency, LIGER employs an intuitive method that aligns object appearances across images. Specifically, the error detector compares objects in successive images, identifying whether two objects should have similar appearance with the command $Modify(object \ in \ V'_i, object \ in \ V_{i-1})$. Subsequently, a locator tool, \ie LISA \citep{lai2024lisa} outputs the masks of the objects according to the object descriptions generated by the error detector. Then the identity-keeping tool \ie  DragonDiffusion \citep{mou2023dragondiffusion} receives the masks and modifies the object appearance in the current image to match the previous image. \\
%Subsequently, a locator tool,\ie, LISA \citep{lai2024lisa} traces the positions of the objects and the identity-keeping tool \ie DragonDiffusion \citep{mou2023dragondiffusion} modifies the object appearance in the current image to match the previous image. 
\textbf{Wrong attribute.} Correct object attributes such as color, shape, and state are crucial for instructions. For instance, considering the tasks of \textit{baking chicken wings}, the model may incorrectly generate cooked chicken wings at the \textit{seasoning the prepared chicken wings} step, where they should be raw. To address this problem, the error detector describes the desired attributes for an object with the instruction $Add(new \  description, object \ in \ V'_i)$. The same locator tool segments the object, then an attribute reformulation tool \ie SD inpainting \cite{rombach2022high} generates an image with modified object attributes according to the object mask. \\
\textbf{Redundant object.} The last type of error is object hallucination, where frozen text-to-image diffusion models sometimes generate irrelevant objects for a step description.
For instance, in Figure \ref{fig:errortype} (b), the image illustrating \textit{preheating the oven} mistakenly includes bread in the pan. \begin{wrapfigure}{r}{0.46\textwidth}
    \vspace{-1.5\intextsep}
    %\hspace{0.1\intextsep}
    \begin{minipage}{0.46\textwidth}
        \begin{algorithm}[H]\small
                \textbf{Input:} Draft Image $V'_i$, Previous Image $V_{i-1}$, \\
                Step Description $S_i, S_{i-1}$, and Task $Q$.\\
                \eIf{$i = 0$}{
                $\sA \gets [\textit{Attribute, Object}]$}
                {$\sA \gets[\textit{Relation, Identity, Attribute, Object}]$}
            \For{$A$ in $\sA$}{
                \eIf{$A$ in $[\textit{Attribute, Object}]$}
                {$error \gets$ Detect($V'_i$, $S_i$, $Q$)}
                {$error \gets$ Detect($V'_i$, $S_i$, $Q$, $S_{i-1}$, $V_{i-1}$)}
                \If{$error$ is detected}{
                    $\hat{V}_i \gets$ Rectify($V'_i$, $S_i$, $Q$)\\
                    $V_i \gets$ Compare($V'_i$, $\hat{V}_i$)\\
                    \textbf{break}}
            }
            \textbf{if} $V_i = \hat{V}_i$ \textbf{then} Refresh($\hat{V}_i$) \textbf{end}\\
            \textbf{Output:} Final Image $V_{i}$,
            \caption{Single Step Self-reflection}
            \label{algorithm}
        \end{algorithm}
    \end{minipage}
    \vspace{-1.5\intextsep}
\end{wrapfigure}The error detector flags the object to be removed in a format of $Remove(object \ in \ V'_i)$, and the locator tool pinpoints the specific region. LIGER opts for the widely used LAMA \citep{suvorov2022resolution} as an object removal tool. The tool removes the corresponding part of the image given the object mask.\\
LIGER evaluates the image across these four aspects iteratively and only modifies the draft image for once. In other words, once an error is detected, the verification procedure halts, and the corresponding editing operation is applied to the draft image. It is also worth noting that the over-consistent and identity inconsistent errors are verified based on two consecutive steps, while wrong attribute and redundant object are conducted as single-image verifications. The execution order of the pipeline is detailed in Algorithm \ref{algorithm}. Consequently, for the draft image of the first step in each task, LIGER only performs attribute modification or object removal.
Having the various tools collaboratively verify the images, LIGER generates illustrative visual instructions for long-horizon tasks with accurate logic in a self-reflection manner.

\subsection{Judgement and Memory Calibration}

The aforementioned tool-based self-reflection generates a revised image $\hat{V}_i$. Yet every rose has its thorn, self-reflection sometimes produces low-quality images or makes incorrect judgments during editing. To stabilize the pipeline predictions and improve robustness, we devise a referee tool to compare the draft image with the revised image. The referee evaluates both the quality and semantic alignment of the images and selects the better one as the final result $V_i$. For more details, refer to the prompt template provided in the appendix. Since LIGER generates images step by step, with visual memory providing visual continuity between steps, any error in the output image $V_i$ impacts the memory and can accumulate in subsequent steps of image generation. To prevent this exposure bias, we propose inversion-guided visual memory calibration to update the memory.\\
\textbf{Inversion-guided visual memory calibration. } As discussed in Section \ref{history}, the visual memory is a set of image feature tokens sampled from the previous generation step $p_{i-1} \in \sR^{M\times C}$. These tokens are saved during the denoising process of the draft image, which exhibits a discrepancy with the features of the revised images. Since the revised image is generated in a post-processing manner, storing the feature tokens alongside the generation process is inapplicable. However, the sampling process can be reversed using DDIM inversion which is formulated as:
\begin{equation}
\vx^{t+1}=\sqrt{{\alpha_{t+1}}/{\alpha_t}} \cdot \vx^t+\sqrt{\alpha_{t+1}}\left(\beta_{t+1}-\beta_{t}\right) \cdot \epsilon_t,
\end{equation}
where $\alpha_t$ is the variance schedule depend on timestep $t$, and the step-wise coefficient is set to $\beta_t = \sqrt{1 /{\alpha_{t}}-1}$. $\epsilon_t$ is the noise predicted by the U-Net according to Eq \ref{eq1}. This allows us to obtain the attention output of the U-Net during the inversion procedure. Therefore, for the revised images, we apply this inversion operation over the same number of timesteps as in the generation procedure, effectively calibrating the visual memories to current image $V_i$ features. Correcting the visual memories prevents accumulated errors affecting subsequent image generation procedures.

\section{Experiments}
\label{exp}
\subsection{Implementation Details}
For the historical textual prompt, the error detector and referee, we use GPT-4O \citep{achiam2023gpt} introduced by OpenAI. The draft image generation uses the SDXL \citep{podell2023sdxl} with a guidance scale of 5 along with the Free-U plugin \citep{si2024freeu}.  The DDIM generation and inversion timesteps are set to 50. In terms of the visual memory, we set the number of the previous step image feature token $M$ to half of the sequence length $N$, in other words, $M = N/2$. For the location tool, we leverage the LISA-7B model \citep{lai2024lisa} to balance the performance and computing resources requirement. All experiments are conducted on a single RTX A6000 GPU.
\begin{figure}[t]
    \centering
    \vspace{-0.5\intextsep}
    \includegraphics[width=\linewidth]{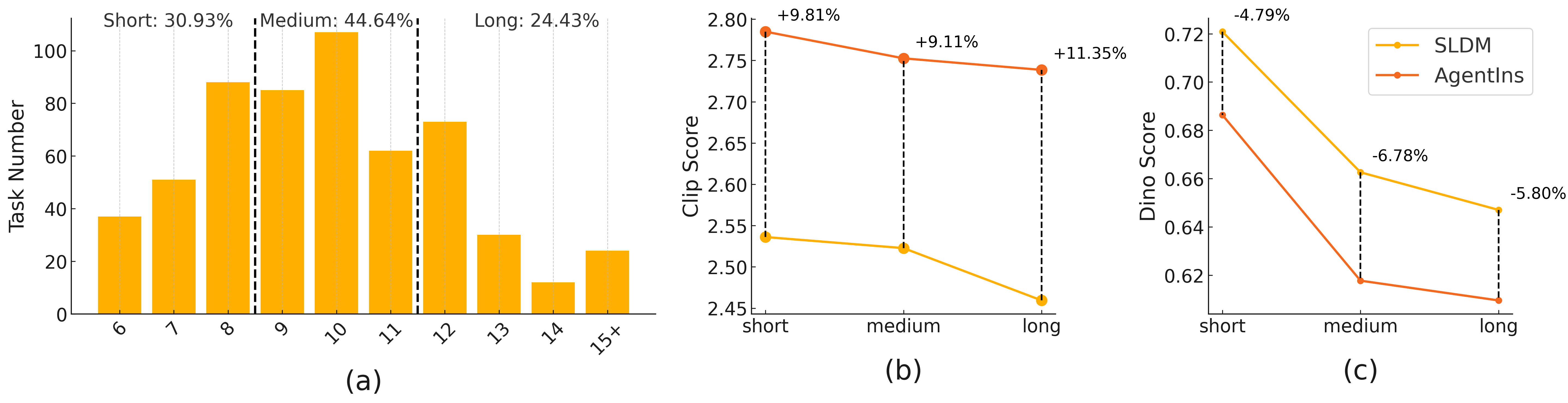}
    \vspace{-1.5\intextsep}
    \caption{Dataset statistics and the influence of the step length of tasks.}
    \vspace{-1\intextsep}
    \label{fig:dataset}
\end{figure}

\subsection{Dataset}
Effective visual instructions for long-horizon tasks should help users quickly understand complex procedures, but evaluating this capability remains challenging. Existing datasets lack appropriate evaluation methods for this aspect. To address this gap, we curate a new textual dataset consisting of 569 long-horizon tasks. These tasks are extracted from different resources including Howto100M \citep{miech2019howto100m}, Youcook2 \citep{zhou2018towards}, and RecipeQA \citep{yagcioglu2018recipeqa}. The tasks focus on the recipe domain, as cooking procedures typically involve strong logical relations between steps and require multiple stages. Specifically, we prompt the GPT4O model with in-context samples to filter out tasks that are hard to illustrate and tasks that are easy to accomplish, \eg \  \textit{How to prepare a family meal for 20 people}. The LLM then outputs step-by-step action descriptions for each task. Unlike existing planning datasets \citep{menon2024generating,lu2023multimodal}, our dataset offers following novel features: \\
\textbf{Long-horizon tasks.} The average number of steps per task is 9.8, with a minimum of 6 steps and a maximum of 17. The detailed distribution is shown in Figure \ref{fig:dataset} (a). We categorize the tasks into three types: short (6-8 steps), medium (9-11 steps), and long (12 or more steps).\\
\textbf{Manual annotations for step logics.} For each task, we ask human annotators to select a pair of consecutive steps with continuous logic and another pair with logically independent steps. Our intuition is that the images corresponding to logically consistent steps should exhibit visual continuity, while the images of locally independent steps should be visually distinct.\\
\textbf{Human-written ground truth descriptions reflecting comprehension.} We introduce a novel annotation for evaluating illustrative images. Since step descriptions often omit details about object attributes, we ask the annotators to write a sentence describing what components should appear in the illustrative image for every step. These sentences reflect the appearance and state of the objects with previous steps information. For example, the step \textit{Arrange the chicken wings on the wire rack} from task \textit{How to bake chicken wings}, one can infer the wings are raw and ready for baking. Therefore, a suitable illustrative expression could be \textit{The raw chicken wings are neatly arranged in a single layer on the wire rack, with the spices and oil giving the skin a glossy, seasoned appearance.} These expressions allow us to evaluate whether the generated images match human expectations of how an illustrative image should look. Annotation examples are provided in the appendix. 
\subsection{Baselines}
To thoroughly evaluate the effectiveness of LIGER and its components, we conduct both quantitative and qualitative comparisons with different baselines including:
(1) \textbf{Frozen SDXL \citep{podell2023sdxl}.} We simply generate visual instructions for the tasks using a frozen SDXL model prompted with the vanilla textual step descriptions. 
(2) \textbf{Frozen SDXL + Visual memory (+V).} The image generation model is provided with the visual memory while the text prompts remain vanilla step descriptions.
(3) \textbf{Frozen SDXL + Historical prompt (+H).} The text prompt for the frozen SDXL model is modified by concatenating the step description and the historical prompt. No visual memories are provided.
(4) \textbf{Frozen SDXL + Visual Memory + Historical prompt (+V+H).} The image generation model is equipped with both visual memory and the historical prompt. This baseline can also be considered LIGER without self-reflection.
(5) \textbf{T2I-Bridge} \citep{lu2023multimodal} uses an LLM to imagine what the image for each step should depict based on the step descriptions. T2I-Bridge represents a type of re-captioning method.
(6) \textbf{Sequential Latent Diffusion Model (SLDM)} \citep{bordalo2024generating} trains a language model to produce coherent captions for the steps of a task and uses a sequential context decoder to establish visual connections between images. Note that the text-to-image generation diffusion model is still frozen in SLDM.

\begin{figure}[t]
    \centering
    \vspace{-1.5\intextsep}
    \includegraphics[width=\linewidth]{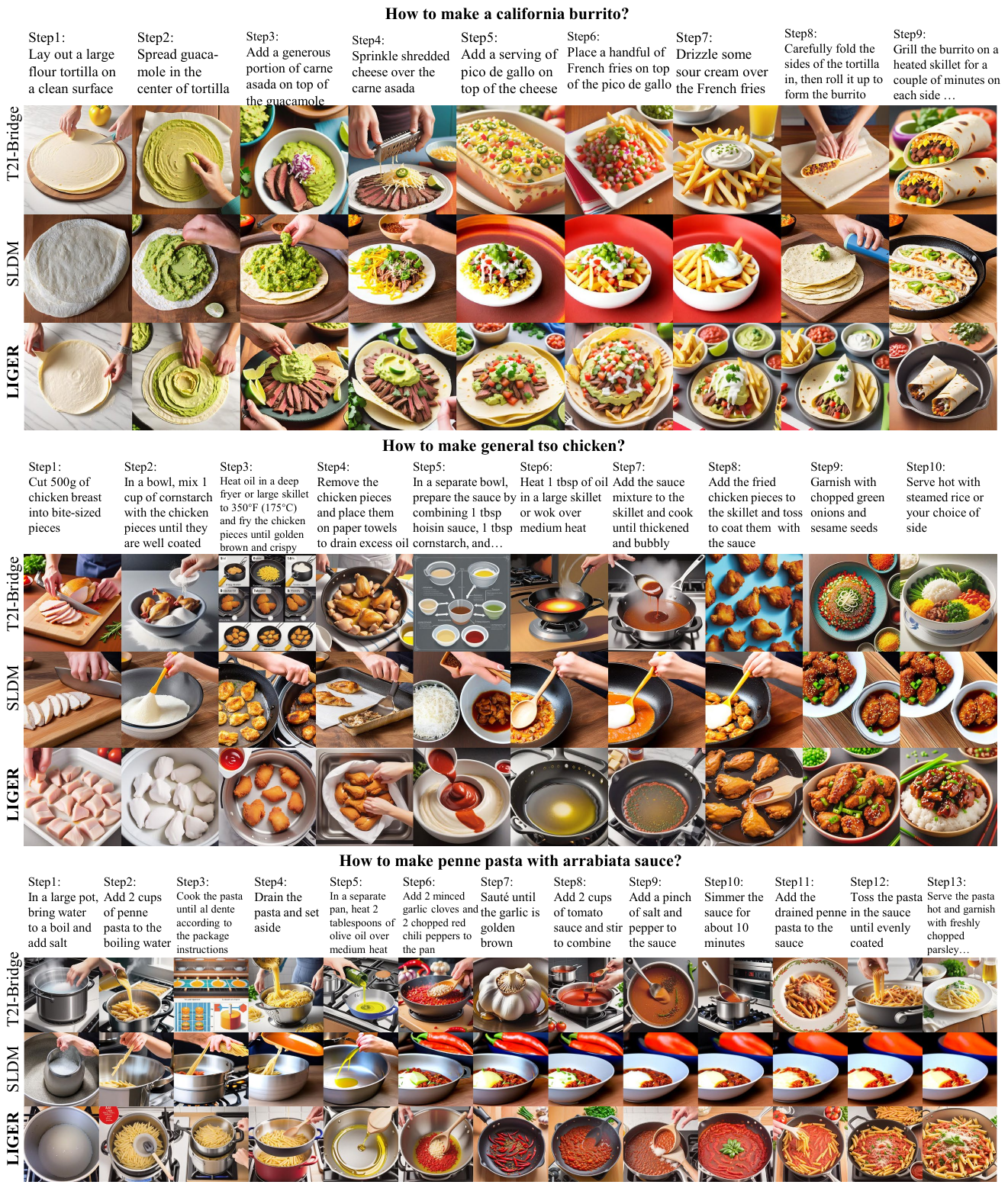}
    \vspace{-1.7\intextsep}
    \caption{Detailed qualitative comparisons on different long-horizon tasks. Zoom in to see details.}
    \vspace{-1.7\intextsep}
    \label{fig:qualitative}
\end{figure}

\subsection{Quantitative Evaluation}
To assess the effectiveness of LIGER, we conduct a detailed quantitative comparison including:
\begin{table}[h]
    \centering
    \resizebox{1.0\linewidth}{!}{
    \begin{tabular}{r|ccc|ccc}
    \toprule
       \multirow{2}{*}{Method} & \multicolumn{3}{c|}{Automatic evaluation} & \multicolumn{3}{c}{GPT evaluation} \\
       
         & CLIP-Score$\uparrow$ & DINO-Score $\downarrow$ & BERT-Score $\uparrow$ & Semantic$\uparrow$ & Logic$\uparrow$ & Illustrative$\uparrow$ \\
        \midrule
        T2I-Bridge  & 2.4350 & 0.8576 & 0.8669 &3.4717 & 2.5843 & 2.5150\\
        SLDM & 2.5054 & 0.6746 & 0.8694 &  3.3634 & 2.7286 & 2.5771 \\
        \textbf{Ours}  & \textbf{2.7555} & \textbf{0.6338} & \textbf{0.8743} & \textbf{4.1141}  & \textbf{3.0595} & \textbf{3.0536} \\
        \bottomrule
    \end{tabular}
    }
    \vspace{-0.5\intextsep}
    \caption{Automatic quantitative evaluation and GPT evaluation results.}
    \vspace{-1.2\intextsep}
    \label{tab:automatic}
\end{table}
\textbf{Automatic evaluation.} We calculate several metrics using pre-trained models. First, we evaluate the semantic alignment between the images and human-annotated ground truth expressions by calculating the CLIP \citep{radford2021learning} similarity. These curated expressions reflect human understanding of each step. Hence a higher CLIP-Score indicates that the images are more relevant to the expressions, implying that the images are more illustrative for human comprehension. 

The second metric tests the logic correctness between consecutive steps. To evaluate image similarity, we use the DINO-v2 \citep{caron2021emerging, oquab2023dinov2} model and calculate the average $l_2$ Distance between the embeddings of the two images for the annotated step pairs. Inspired by the Signal-to-Noise Ratio formulation, we define the DINO-Score $D_s$ as the $l_2$ distance between coherent steps divided by the $l_2$ distance between independent steps which can be expressed as $D_s = l_2^p/l_2^n$. This metric evaluates the ability to generate consistent images for logically coherent steps and distinct images for unrelated steps. A lower DINO-Score indicates higher logical accuracy. 

The last metric evaluates the method performance in a modality-transfer test. Our intuition is that illustrative visual instruction should help people summarize or describe the steps in text. Therefore, we transfer the images back into text and measure the textual similarity with the annotated descriptions. Specifically, we adopt the widely-used BLIP-2 \citep{li2023blip} model to generate captions for images, then calculate the BERT-Score \citep{zhang2019bertscore} between the captions and descriptions. A higher BERT-Score represents the image is more illustrative. The results shown in Table \ref{tab:automatic} demonstrate that LIGER significantly outperforms the baseline methods.
\begin{wrapfigure}{r}{0.65\textwidth}
    \vspace{-1\intextsep}
    %\hspace{0.1\intextsep}
    \centering
    %\vspace{-1.5\intextsep}
    \includegraphics[width=\linewidth]{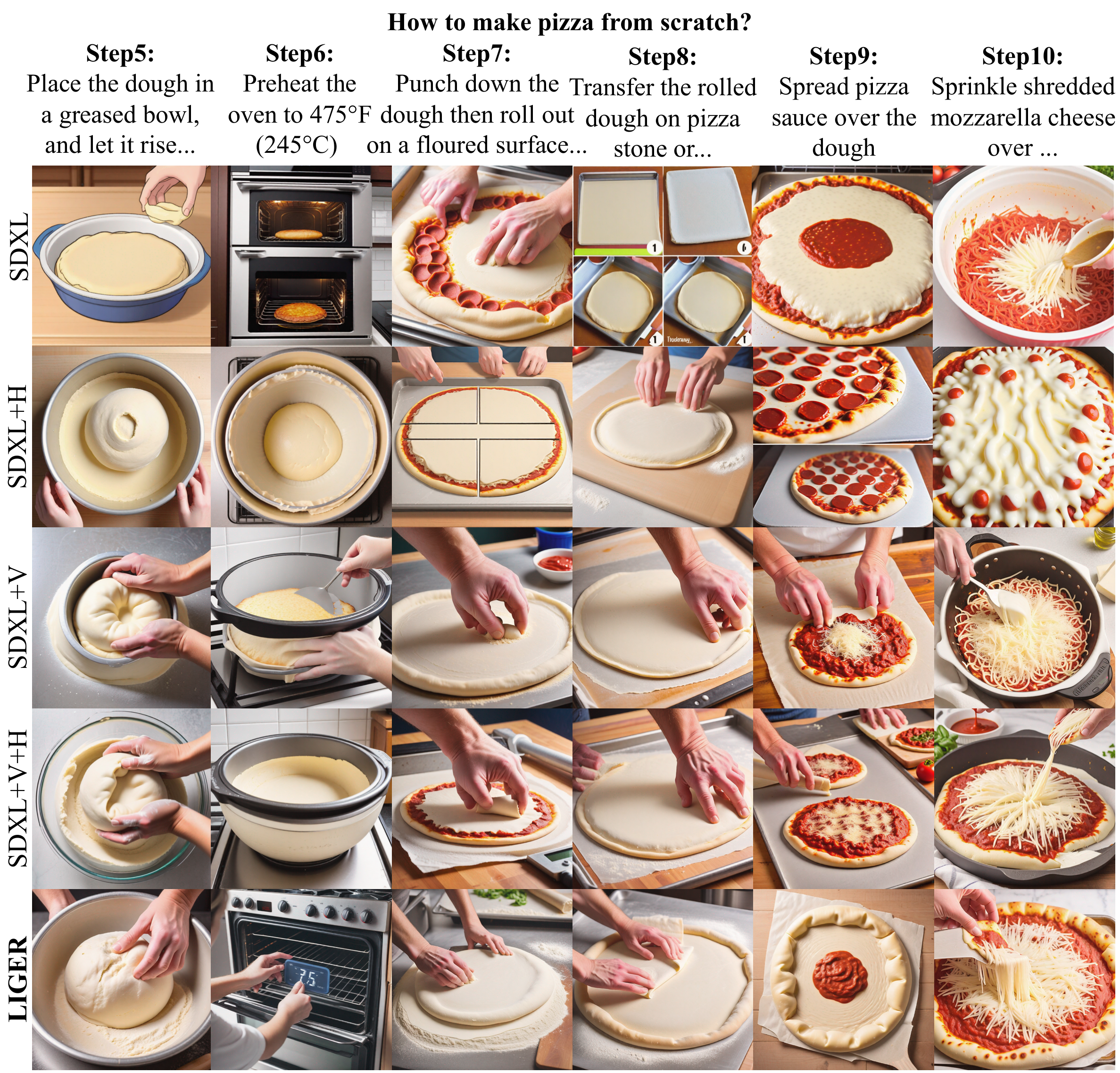}
    \vspace{-1.5\intextsep}
    \caption{Qualitative ablation on different components. }
    \vspace{-1\intextsep}
    \label{fig:ablation}
    %\vspace{-1.5\intextsep}
\end{wrapfigure}
\textbf{GPT evaluation.} We further harness the advanced logical reasoning and multi-modal perception ability of MLLMs to evaluate the methods. Specifically, we prompt the GPT4O model to rate how well each individual image aligns with its corresponding description. Then we input the entire image series to the MLLM and ask it to rate whether the image series is illustrative with correct logics. The rating ranges from 1 to 5, where 1 represents low quality and 5 indicates perfect quality. The results are shown in Table \ref{tab:automatic}, and the prompt templates are attached in the appendix.\\
\textbf{User study.} We invite 20 participants for the user study, with each person asked to select the best generation results for 15 tasks. Participants rate aspects including semantic alignment, logical correctness, and task illustration. Results in Table \ref{tab:userstudy} show that LIGER generates visual instructions that better match user preferences while maintaining semantic alignment and logic accuracy.

\subsection{Qualitative Comparisons} 
The overall qualitative comparisons between LIGER and baseline methods are shown in Figure \ref{fig:qualitative}. We provide a detailed comparison of LIGER with two prior works, namely T2I-Bridge and SLDM. For the task \textit{How to make a California burrito}, both T2I-Bridge and SLDM overlook that the seasoning and ingredients are added to the tortilla in Steps 3 to 7. In contrast, LIGER clearly illustrates the progressive process of adding different ingredients. Additionally, LIGER correctly visualizes the burrito being wrapped and heated in a skillet. For the task \textit{How to make general tso chicken}, LIGER presents a smooth sequence, showing the process of frying the chicken pieces, making the sauce, combining sauce with chicken, and serving with rice. In comparison, SLDM omits the chicken pieces in Step 2 and incorrectly shows the finished dish in Step 5. T2I-Bridge lacks visual continuity, making it hard to comprehend.
To further demonstrate the effectiveness of LIGER in long-horizon tasks, we visualize the results for the task \textit{How to make penne pasta with arrabbiata sauce} consisting of 13 steps. SLDM shows an over-consistent process during cooking, while T2I-Bridge generates distinct images. In contrast, LIGER accurately illustrates the procedure.
\subsection{Ablation Study} 
\textbf{Effectiveness of different components.} We provide both qualitative and quantitative comparisons in Figure \ref{fig:ablation} and Table \ref{tab:ablation}.
Results show that adding historical prompts and visual memory both improve the alignment between image and text semantics while also increasing logical accuracy. Additionally, these two components complement each other. When self-reflection is introduced, we observe a performance gain of +0.04 in CLIP-Score, a reduction of -0.112 in DINO-Score, and an improvement of +0.002 in BERT-Score, demonstrating the importance of self-reflection. In Figure \ref{fig:ablation}, we observe that self-reflection correctly identifies which steps should be visually coherent and which steps should be distinct. Moreover, LIGER effectively shows the process of transforming pizza dough into a raw pizza. Essentially, the historical prompt and visual memory enhance visual continuity, while self-reflection aligns the images with human comprehension.

We further provide an example to highlight the importance of visual memory calibration in Figure \ref{fig:inversion}. For Step of \textit{season the steak}, the steak should be raw, yet the draft image incorrectly shows a cooked appearance. After correcting the attribute, the subsequent step should also depict the steak as raw since the description does not indicate a state change. Without memory calibration, the steak in the next step still appears cooked, but with calibration, the steak is correctly shown in a raw state.

\vspace{-0.5\intextsep}
\begin{minipage}[t]{0.54\linewidth}
    \vspace{0\linewidth}
    \centering
    \resizebox{1.0\linewidth}{!}{
    \begin{tabular}{r|ccc}
    \toprule
            & CLIP-Score $\uparrow$ & DINO-Score $\downarrow$ & BERT-Score $\uparrow$ \\
        \midrule
       SDXL & 2.5837 & 0.8516 & 0.8699\\
       SDXL+V & 2.6251 & 0.8239 & 0.8719\\
       SDXL+H & 2.6842 & 0.8224 & 0.8707\\
       SDXL+V+H & 2.7168 & 0.7459 & 0.8721\\
       \textbf{Ours} & \textbf{2.7555} & \textbf{0.6338} & \textbf{0.8743}\\
       %  & CLIP \uparrow & DINO \downarrow & BERT \uparrow \\
       %  \midrule
       % SDXL & 2.58420 & 0.8572 & 0.8699\\
       % +L & 2.62562 & 0.8301 & 0.8719\\
       % +H & 2.68485 & 0.8323 & 0.8707\\
       % +L+H & 2.71747 & 0.7574 & 0.8722\\
       % \textbf{Ours} & \textbf{2.75611} & \textbf{0.6486} & \textbf{0.8743}\\
        \bottomrule
    \end{tabular}
    }
    
    \captionof{table}{Ablation on different components of LIGER.} 
    \label{tab:ablation}
    \vspace{0.2\intextsep}
     \resizebox{0.8\linewidth}{!}{
     
    \begin{tabular}{cccc}
        \toprule
        Method & Semantic & Logic & Illustrative \\
        \midrule
        T2I-Bridge & 24\% & 18.3\% & 22.3\% \\
        SLDM & 11.7\% & 21\% & 9.3\% \\
        \textbf{Ours} & \textbf{64.3\%} & \textbf{60.7\%} & \textbf{68.3\%} \\
        \bottomrule
    \end{tabular}}
    \captionof{table}{User study on image-text semantic matching, logic continuity and illustrative.} 
    \label{tab:userstudy}
\end{minipage}%
\hfill
\begin{minipage}[t]{0.44\linewidth}
\vspace{-0.4\intextsep}
\vspace{0\linewidth}
    \centering
    \includegraphics[width=\linewidth]{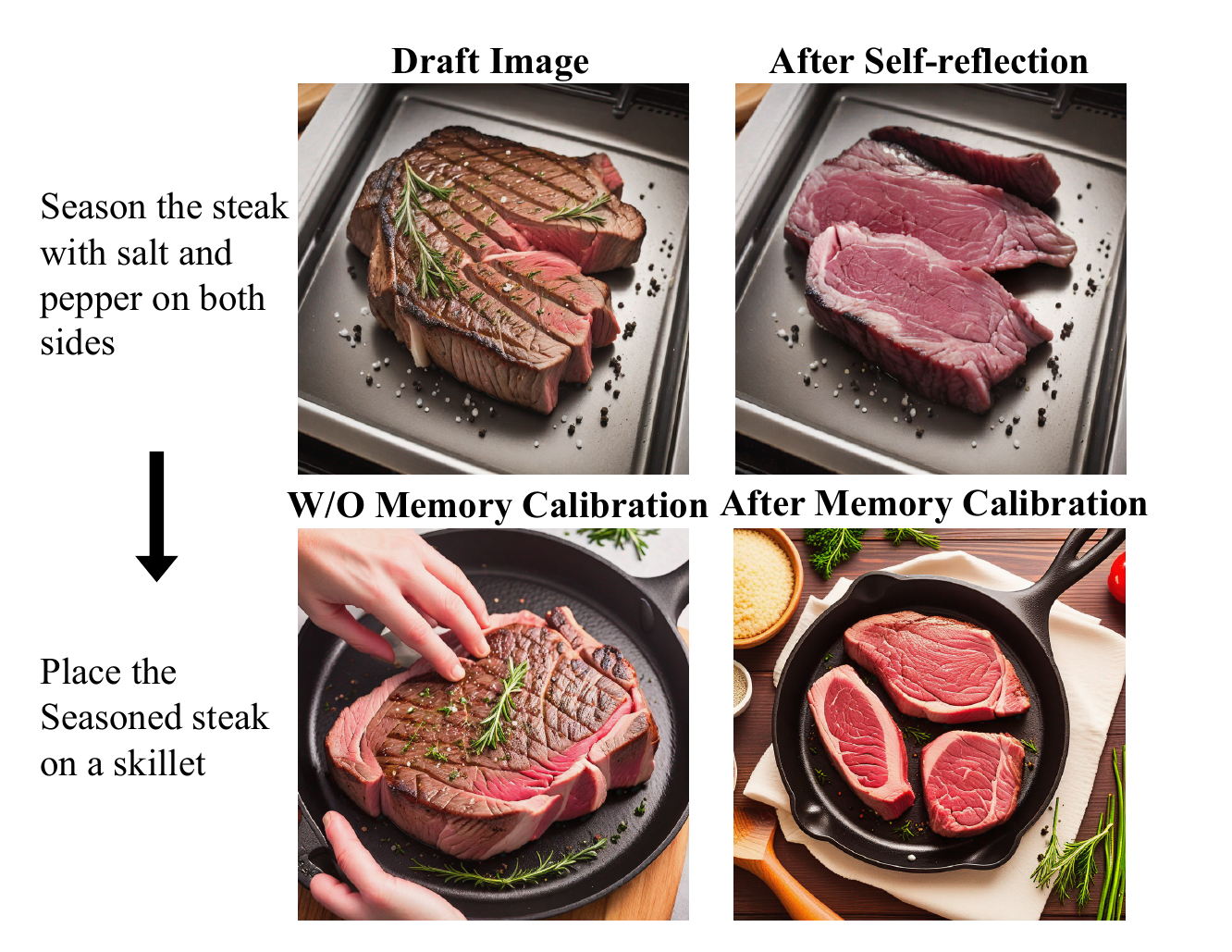} 
    
    \captionof{figure}{Example of visual memory calibration.}
    \label{fig:inversion}
\end{minipage}

\textbf{Influence of task step length.} In Figure \ref{fig:dataset} (b) and (c), we present the CLIP-Score and DINO-Score for tasks of varying lengths, comparing LIGER with SLDM. As the number of task steps increases, the CLIP-Score of SLDM decreases significantly, while LIGER maintains stable performance. Additionally, the relative improvement in DINO-Score increases for medium and long tasks, indicating LIGER is robust to long-horizon tasks.
\section{Conclusion}
In this paper, we propose LIGER, the first training-free framework for long-horizon visual instruction generation. LIGER first leverages historical prompts and visual memory to generate draft images step-by-step, enhancing continuity between images in long-horizon tasks. The tool-based self-reflection rectifies four types of errors in the draft images including over-consistent, identity inconsistent, wrong attributes, and object redundant. LIGER also deploys inversion-guided visual memory calibration to prevent error accumulation in the sequential image generation procedure. We also curate a new benchmark testing the alignment of generation results with human comprehension. We hope this work inspires future research on instruction generation.  

\section{Acknowledgements}
This work is supported by the National Natural Science Foundation of China (U2336212). This work is also supported in part by ``Pioneer" and ``Leading Goose" R\&D Program of Zhejiang (No.2024C01142). We are grateful for the user study participants. This work was partially supported by ZJU Kunpeng\&Ascend Center of Excellence. We also thank Dr Xiao Pan for discussing about the paper writing.

\bibliography{iclr2025_conference}
\bibliographystyle{iclr2025_conference}

\appendix
\section{Appendix}
\subsection{Additional Results}
\begin{figure}[t]
    \centering
    \vspace{-1.5\intextsep}
    \includegraphics[width=\linewidth]{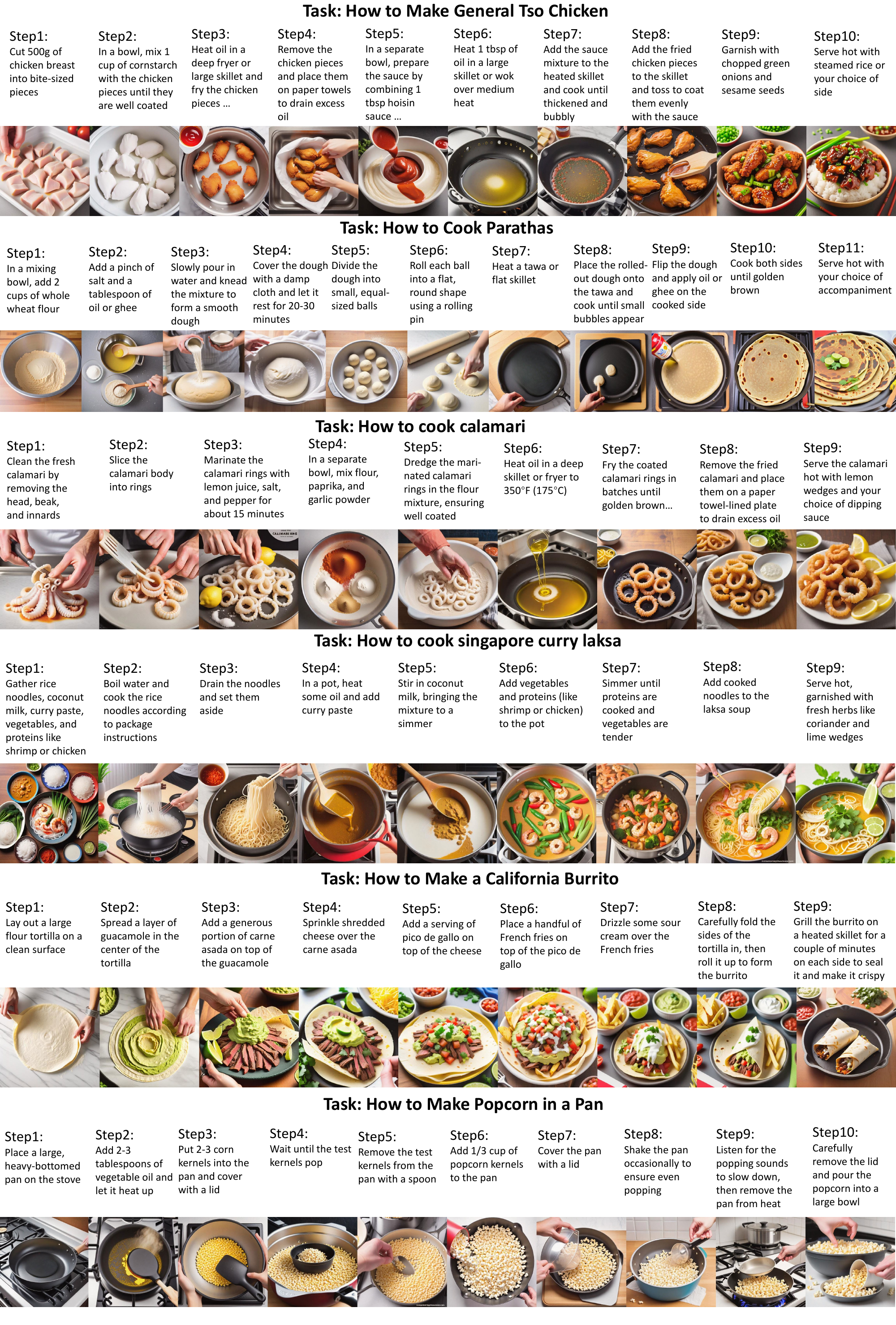}
    \vspace{-0.5\intextsep}
    \caption{Additional qualitative results generated by LIGER. Zoom in to see the detail.}
    \label{fig:appendix}
\end{figure}

\begin{figure}[t]
    \centering
    \vspace{-1.5\intextsep}
    \includegraphics[width=\linewidth]{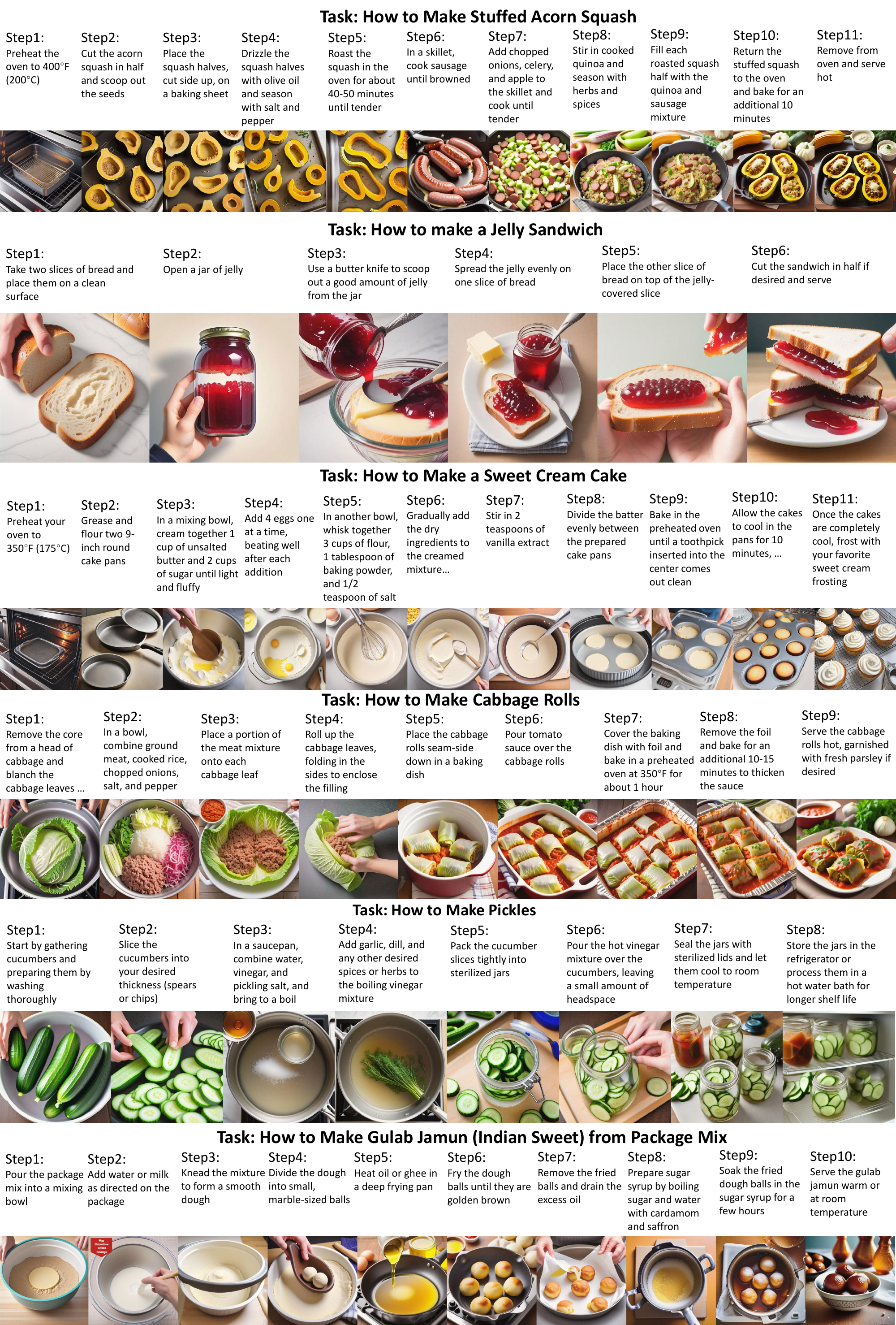}
    \vspace{-0.5\intextsep}
    \caption{Additional qualitative results generated by LIGER. Zoom in to see the detail.}
    \label{fig:appendix1}
\end{figure}

Additional qualitative results are shown in Figure \ref{fig:appendix} and Figure \ref{fig:appendix1}.

\end{document}

%% file: math_commands.tex
\usepackage{amsmath,amsfonts,bm}

\def\eqref#1{equation~\ref{#1}}
\def\1{\bm{1}}

\def\vc{{\bm{c}}}

\def\vp{{\bm{p}}}

\def\vx{{\bm{x}}}

\def\vz{{\bm{z}}}

\DeclareMathAlphabet{\mathsfit}{\encodingdefault}{\sfdefault}{m}{sl}
\SetMathAlphabet{\mathsfit}{bold}{\encodingdefault}{\sfdefault}{bx}{n}

\def\sA{{\mathbb{A}}}

\def\sR{{\mathbb{R}}}
\def\sS{{\mathbb{S}}}

\def\sV{{\mathbb{V}}}